\documentclass{article}
\usepackage{spconf,amsmath,amssymb,graphicx,booktabs,multirow}

\title{SYMMETRY-AWARE LIKELIHOOD-ORBIT AGGREGATION FOR SELECTIVE LEFT--RIGHT CLAIM VERIFICATION}

\name{Zhouzhi Xiong$^{1,*}$, Chuxi Zhang$^{4,*}$, Weizhen He$^{2}$,
Yi Chen$^{3}$, Qi Li$^{2}$, Donglian Qi$^{2}$}
\address{$^{1}$Polytechnic Institute, Zhejiang University, Hangzhou, China\\
$^{2}$College of Electrical Engineering, Zhejiang University, Hangzhou, China\\
$^{3}$Ocean College, Zhejiang University, China\\
$^{4}$College of Artificial Intelligence, Zhejiang University, Hangzhou, China\\
$^{*}$Equal contribution}

\begin{document}
\maketitle

\begin{abstract}
Frozen vision--language models (VLMs) remain unreliable on fine-grained
left--right claims, and raw claim likelihoods need not reliably rank
verification errors. After a horizontal-reflection intervention is fixed, how should its induced
likelihood measurements be combined into a selective verification signal? We
introduce \emph{Relation-Orbit}, a closed-form contrast with no learned fusion
parameters that
assigns eight normalized likelihoods to query-supporting and counterfactual
roles determined by reflection, inverse relation, and entity exchange. A
claim is asserted only when the signed contrast exceeds a threshold selected
on held-out data using pointwise Clopper--Pearson upper confidence bounds. On VSR and GQA across four frozen VLMs, Relation-Orbit yields
higher mean test coverage at a 10\% selective-risk calibration target than
an all-eight Orbit-Max baseline in all eight dataset--backbone settings; gains over a nearly abstain-all
one-sided intervention score are reported separately. A separate LLaVA-1.5/COCO evaluation, reduced-orbit controls, and a
two-sided partition diagnostic further characterize the structural advantage.
\end{abstract}

\begin{keywords}
vision--language models, selective prediction, claim verification, spatial
relations, likelihood aggregation
\end{keywords}

\section{Introduction}
\label{sec:introduction}

Vision--language models can identify two objects yet reverse which one lies
to the left of the other~\cite{liu2023vsr,kamath2023whatsup,yang2019spatialsense,
thrush2022winoground}. Such errors
are problematic whenever downstream decisions rely on spatial claims.
Improving average accuracy alone is
insufficient in these settings: the system should also abstain when the
available evidence is insufficient to meet an explicit error target.

Selective prediction turns model evidence into a decision signal and asserts a
claim only when the signal exceeds a
calibrated threshold~\cite{chow1970reject,geifman2019selectivenet,
angelopoulos2024conformalrisk}.
This separation matters because modern neural confidence can be
miscalibrated even when average accuracy is high~\cite{guo2017calibration}.
For left--right questions, we adopt the horizontal-reflection intervention
studied by BCEA~\cite{xu2026bcea}, which compares a relational claim across
the original and reflected views. Our question begins after this intervention
has been fixed: a one-sided contrast uses only one textual formulation and
leaves the inverse-relation and entity-exchange structure among the induced
likelihoods unused.

We ask a complementary question: \emph{given a fixed intervention, how
should all likelihood observations induced by it be organized?} A left
claim, its inverse, and their entity-swapped formulations are not unrelated
prompts. Their truth values transform predictably under horizontal
reflection. Relation-Orbit encodes these roles in a closed-form signal.
This distinction is important: our contribution is not merely to ``use eight
likelihoods,'' but to organize them by the known relation symmetry. We compare the resulting score with one-sided scoring, an all-eight
Orbit-Max control, and reduced-orbit contrasts in the main assertion
experiments. A separate structural diagnostic evaluates it against
alternative four-versus-four partitions.

Related work addresses different stages of reliable VLM reasoning. ReCoVERR
retrieves and verifies external evidence for uncertain answers~\cite{srinivasan2024recoverr},
while Budgeted Conformal Evidence Acquisition (BCEA) selects budgeted evidence
and recalibrates after acquisition~\cite{xu2026bcea}. We begin after a fixed
intervention is selected and ask how to aggregate every likelihood it induces. SEER constructs query-specific grounded views with explicit entity roles and
reciprocal consistency~\cite{liu2026seer}; Relation-Orbit instead uses a fixed
reflection and organizes four linked formulations across two views into
supporting and counterfactual orbits. SAGE enforces geometric--linguistic
duality consistency across transformed inputs during VLM
post-training~\cite{liu2026sage}; in contrast, Relation-Orbit keeps the VLM
frozen and aggregates the complete transformation-induced likelihood orbit
at inference time for selective verification.
Black-box consistency~\cite{khan2024consistency} and visual contrastive decoding~\cite{leng2024vcd} use repeated responses or
perturbed visual evidence to assess reliability. A detector--geometry trust
predictor estimates reliability from detection and geometry~\cite{imran2026trust}.
\emph{Grounding Isn't Knowing} separately analyzes how localization relates to
spatial reasoning~\cite{liu2026grounding}. Our score has no learned parameters
but requires labeled held-out calibration under the common protocol.

Our contributions are threefold. First, we define a closed-form
Relation-Orbit score for left--right claim verification in frozen
VLMs. Second, we establish its symmetry and nuisance-cancellation properties.
Third, cross-protocol experiments, matched-budget and reduced-orbit controls,
and a label-independent structure audit test whether the complete orbit
provides value beyond likelihood count alone. Pointwise CP-UCB serves as
the held-out calibration protocol; we report realized test risk without
claiming a new end-to-end guarantee. We do not claim general spatial reasoning
or a universal visual involution.

\section{Relation-Orbit Aggregation}
\label{sec:method}

\subsection{Problem setting}

Let $x$ be an image, $A$ and $B$ the queried entities, and
$r\in\{\textsc{left},\textsc{right}\}$. The task is selective verification
of the query claim $c_1=(A,r,B)$: the system either asserts $c_1$ or abstains.
Let $y\in\{0,1\}$ indicate whether $c_1$ is true.
We assume a frozen VLM that exposes token likelihoods and a fixed horizontal
reflection $T$. Let $\iota(r)$ be the inverse relation. From entity exchange
and relation inversion we form
\begin{align}
c_1&=(A,r,B), & c_2&=(B,\iota(r),A),\nonumber\\
c_3&=(A,\iota(r),B), & c_4&=(B,r,A).
\label{eq:claims}
\end{align}
Claims $c_1,c_2$ are equivalent; $c_3,c_4$ express the opposite relation.
For view $v\in\{x,T(x)\}$, let $L_{v,j}$ be the length-normalized log
likelihood of the claim span in a fixed prompt containing $c_j$.

\begin{figure*}[t]
  \centering
  \includegraphics[width=0.99\textwidth]{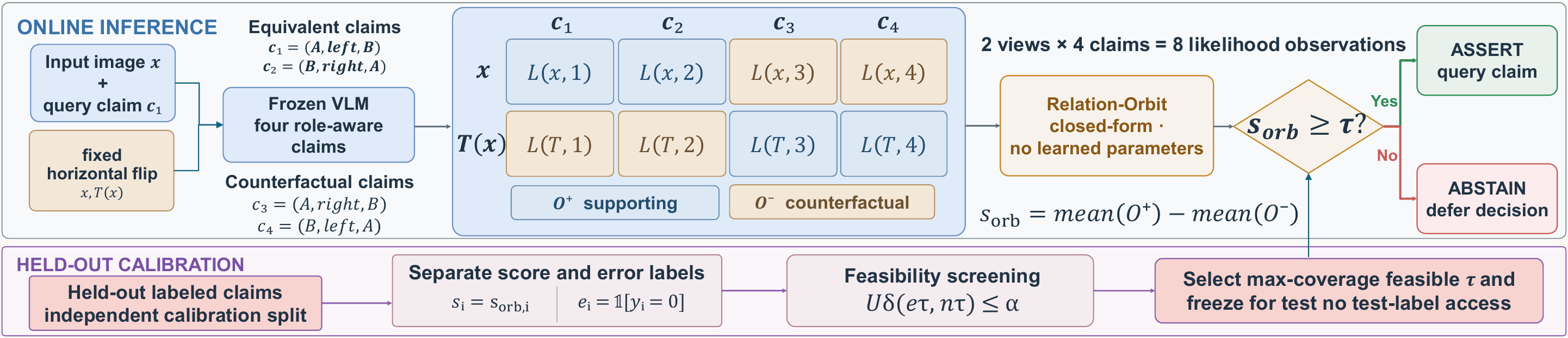}
  \caption{Relation-Orbit organizes eight likelihood observations induced by
horizontal reflection and four role-aware claims into supporting
($\mathcal O^+$) and counterfactual ($\mathcal O^-$) orbits. Their
closed-form contrast is calibrated on held-out labeled claims for selective
assertion; no VLM or fusion parameters are learned.}
  \label{fig:method}
\end{figure*}

\subsection{Closed-form signal construction}

Figure~\ref{fig:method} summarizes the complete inference and calibration
pipeline. Reflection swaps left and right while preserving entity identity. Thus the
supporting observations are $c_1,c_2$ on $x$ and $c_3,c_4$ on $T(x)$; the
remaining observations support the counterfactual:
\begin{align}
\mathcal O^+&=\{L_{x,1},L_{x,2},L_{T,3},L_{T,4}\},\nonumber\\
\mathcal O^-&=\{L_{x,3},L_{x,4},L_{T,1},L_{T,2}\}.
\label{eq:orbits}
\end{align}
Relation-Orbit is the parameter-free contrast
\begin{equation}
s_{\rm orb}=\frac{1}{4}\sum_{z\in\mathcal O^+}z-
             \frac{1}{4}\sum_{z\in\mathcal O^-}z.
\label{eq:score}
\end{equation}
Larger values provide stronger evidence for the query claim; negative values
favor its counterfactual. No VLM weights or fusion parameters are learned.

\textbf{Proposition 1.} The Relation-Orbit contrast is invariant to joint
entity exchange and relation inversion, antisymmetric to relation inversion
alone, and invariant under $L_{v,j}\mapsto L_{v,j}+b_v+a_j$, where $b_v$ and
$a_j$ are view- and formulation-specific offsets. These properties concern
the constructed score and do not assume that the VLM itself is equivariant.

\textbf{Proof sketch.}
Joint entity exchange and relation inversion maps
$c_1\leftrightarrow c_2$ and $c_3\leftrightarrow c_4$, reindexing terms within
each orbit. Relation inversion alone maps $c_1\leftrightarrow c_3$ and
$c_2\leftrightarrow c_4$, swapping $\mathcal O^+$ and $\mathcal O^-$ and
negating the score. A view offset appears twice with each sign within that
view; a formulation offset appears once with each sign across views, so both
cancel. These properties follow from the prescribed orbit coefficients and
do not require the underlying VLM to be equivariant.

\subsection{Selective-risk calibration}

For calibration claim $i$, let $s_i=s_{{\rm orb},i}$ and let
$e_i=\mathbf{1}[y_i=0]$ denote the false-assertion indicator for the
query-assertion action; the label is not part of the score. For threshold $\tau$, assert the query claim when $s_i\ge\tau$ and
otherwise abstain. On the calibration set,
$n_\tau=\sum_i\mathbf{1}[s_i\ge\tau]$ is the asserted count and
$e_\tau=\sum_i\mathbf{1}[s_i\ge\tau]e_i$ is the number of false assertions.
For $n_\tau>0$, we compute the one-sided Clopper--Pearson upper confidence
bound (CP-UCB)~\cite{clopper1934}
\begin{equation}
U_\delta(e_\tau,n_\tau)=
\operatorname{Beta}^{-1}(1-\delta;e_\tau+1,n_\tau-e_\tau),
\label{eq:cp}
\end{equation}
with $U_\delta=1$ when every asserted calibration claim is incorrect.
Candidates are the distinct finite calibration scores. Among those satisfying
$U_\delta\le\alpha$, we choose the
one with maximum calibration coverage and freeze it for the test split. If no
nonempty acceptance set is feasible, we set $\tau=\infty$ and abstain on the
entire test split; its coverage and recorded risk are both zero by convention.
Coverage is $\Pr(s_{\rm orb}\ge\tau)$; selective risk is
$\Pr(y=0\mid s_{\rm orb}\ge\tau)$.
We use $\delta=0.10$ throughout. Because the implementation scans multiple
candidate thresholds using pointwise bounds, we treat CP-UCB as a calibration
protocol and assess its realized behavior on held-out test partitions; we do
not claim a simultaneous or end-to-end finite-sample
guarantee~\cite{angelopoulos2024conformalrisk}.

Calibration uses $(s_i,e_i)$ but never inserts $e_i$ into the score.
Test labels are used only for retrospective reporting, and the score is not
a calibrated probability.

\begin{table*}[t]
\centering
\caption{Paired test-coverage gain (pp) of Relation-Orbit at
$\alpha=0.10$ over 300 grouped splits; brackets are empirical paired
95\% bootstrap intervals. Orbit-Max uses the same eight likelihoods and
the same $\mathcal O^+/\mathcal O^-$ membership, differing only in
within-orbit pooling.}
\label{tab:main}
\setlength{\tabcolsep}{4.1pt}
\begin{tabular}{llcccc}
\toprule
Dataset & Comparator & Qwen-3B & Qwen-7B & InternVL-2B & InternVL-8B \\
\midrule
\multirow{2}{*}{VSR}
 & One-sided & $14.09\,[13.06,15.20]$ & $5.82\,[5.10,6.63]$ & $9.64\,[8.51,10.82]$ & $13.36\,[12.52,14.20]$ \\
 & Orbit-Max & $2.39\,[1.46,3.35]$ & $1.18\,[0.41,1.91]$ & $5.70\,[4.74,6.73]$ & $3.61\,[2.79,4.44]$ \\
\midrule
\multirow{2}{*}{GQA}
 & One-sided & $12.15\,[11.71,12.66]$ & $1.91\,[1.60,2.23]$ & $3.86\,[3.33,4.39]$ & $5.55\,[5.01,6.06]$ \\
 & Orbit-Max & $8.73\,[8.15,9.34]$ & $1.69\,[1.37,2.00]$ & $3.05\,[2.55,3.63]$ & $5.41\,[4.88,5.92]$ \\
\bottomrule
\end{tabular}
\end{table*}

\section{Experiments}
\label{sec:experiments}

\subsection{Protocol and baselines}

We evaluate balanced left--right subsets of Visual Spatial Reasoning
(VSR)~\cite{liu2023vsr} and
GQA~\cite{hudson2019gqa}. VSR contains 1,186 image groups and 2,372 paired true
and inverse claims; GQA contains 1,500 groups and 3,000 claims. We group by
image, preventing paired claims or orbit observations from crossing partitions.
Each of 300 seeded
grouped splits allocates $40\%/30\%/30\%$ of the images to
train/calibration/test. Closed-form methods use no training data; the training
partition only matches the frozen evaluation pipeline. The
four frozen backbones are Qwen2.5-VL-3B/7B~\cite{bai2025qwen25vl} and
InternVL3-2B/8B~\cite{zhu2025internvl3}. All methods share prompts,
token-normalized likelihoods, and the same threshold-search procedure, with
$\alpha=0.10$ and $\delta=0.10$. We report paired mean coverage differences
(pp) with empirical $95\%$ bootstrap intervals over reused split-level
differences.

Claims use a fixed prompt wrapper and mean log likelihood over claim tokens.
Reflection is deterministic; image variants remain in one split, thresholds
use calibration labels only, and no VLM or fusion parameters are trained. Thus
the comparison changes the signal construction, not the backbone.

The one-sided baseline uses
$s_{\rm one}=L_{x,1}-L_{T,1}$, matching BCEA's horizontal-reflection
intervention but not its complete method.
Our main matched-budget control is
\[
s_{\max}=\max \mathcal O^+ - \max \mathcal O^- ,
\tag{5}
\]
which uses the same eight likelihoods and the same orbit membership but
changes mean pooling to max pooling.

We also evaluate three reduced-orbit controls:
$s_{\rm ctr}=L_{x,1}-L_{x,3}$,
$s_{\rm sym}=\frac12[(L_{x,1}-L_{x,3})+(L_{T,3}-L_{T,1})]$, and
$s_{\rm text}=\frac12[(L_{x,1}+L_{x,2})-(L_{x,3}+L_{x,4})]$.
View-Mean is retained only as a role-agnostic sanity control: it assigns
identical scores to paired true/inverse queries and has zero feasible
coverage in all eight settings.

\subsection{Main results}

Table~\ref{tab:main} shows that Relation-Orbit improves over both one-sided
scoring and Orbit-Max in all eight dataset--backbone settings. The smallest
gain over one-sided is $+1.91$ pp on GQA/Qwen-7B, while the smallest gain
over Orbit-Max is $+1.18$ pp on VSR/Qwen-7B; all paired intervals remain
above zero. One-sided coverage is near zero because nonempty feasible
thresholds rarely exist, making Orbit-Max the more informative matched-budget
comparison.

Against reduced-orbit controls, Relation-Orbit improves in $8/8$ settings
over $s_{\rm ctr}$ and $7/8$ over both $s_{\rm sym}$ and $s_{\rm text}$.
The only significant reversal is VSR/InternVL3-2B:
$-3.10$ pp versus $s_{\rm sym}$ (95\% CI $[-4.03,-2.12]$) and
$-1.26$ pp versus $s_{\rm text}$ ($[-2.11,-0.40]$).
Thus the complete orbit is systematically, but not uniformly, preferable.

\begin{figure*}[!t]
  \centering
  \includegraphics[width=0.80\textwidth]{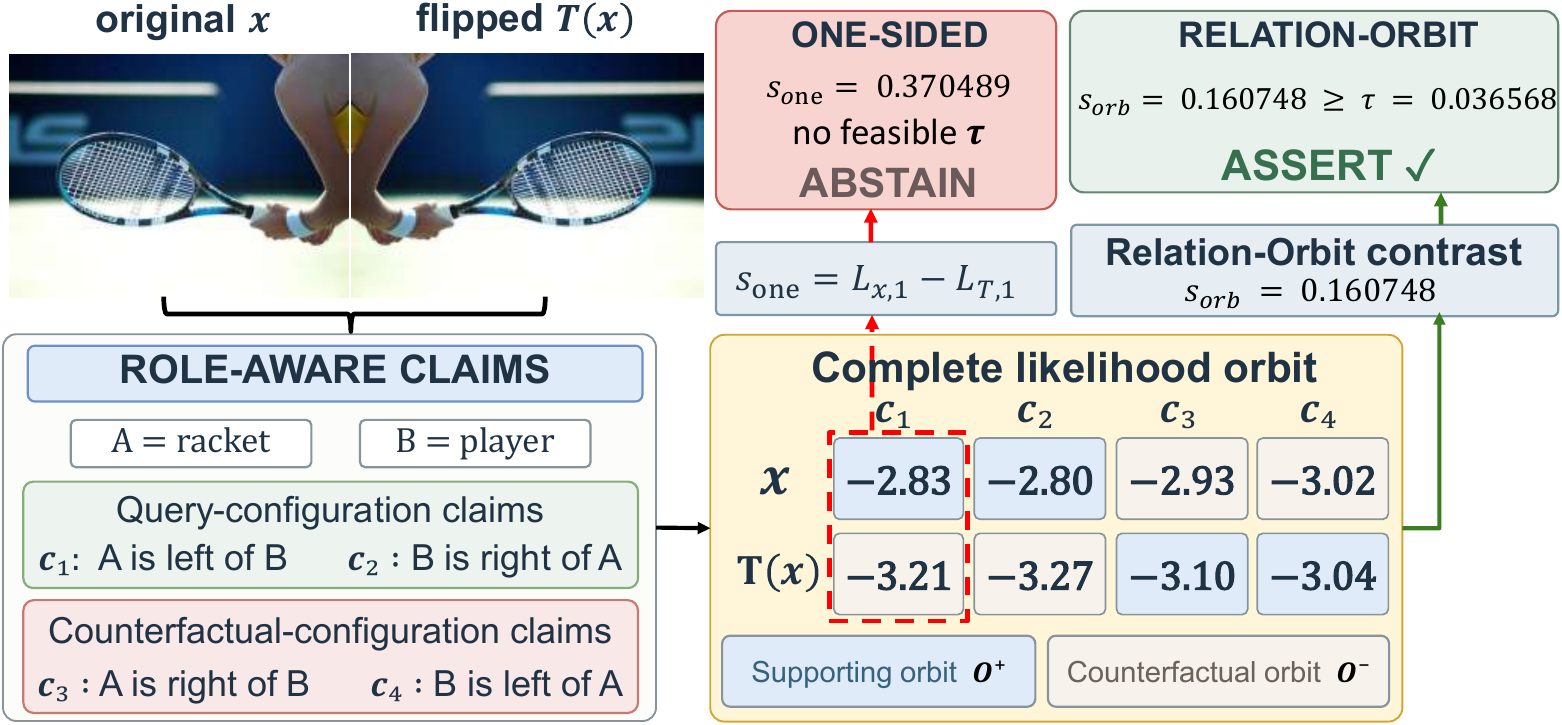}
  \caption{GQA example ($A$: racket; $B$: player). Rows are $x,T(x)$; blue/ochre
  denote $\mathcal O^+/\mathcal O^-$. Relation-Orbit uses all eight cells;
  one-sided uses only the $c_1$ pair. Cell values are rounded for display;
  both $s_{\rm one}$ and $s_{\rm orb}$ are computed from full-precision
  likelihoods. Held-out calibration
  (Qwen2.5-VL-3B, repeat 164, image 2409694) gives the shown orbit threshold
  and no feasible one-sided threshold; its abstention is not a claim-level error.}
  \label{fig:gqa-case}
\end{figure*}

Under the separate BCEA-aligned LLaVA-1.5/COCO
protocol~\cite{lin2014coco,liu2024llava15}, Relation-Orbit gains $34.39$ pp
over one-sided and $13.08$ pp over the matched-input control (paired $95\%$
CIs: $[34.34,34.45]$ and $[13.00,13.16]$).

\begin{table}[t]
\centering
\caption{Absolute Relation-Orbit operating points (\%) over 300 splits.
$C/R$: coverage/recorded selective risk; Q3/Q7 and I2/I8 abbreviate
Qwen2.5-VL-3B/7B and InternVL3-2B/8B.
Empty test acceptance sets are assigned zero recorded risk by convention.}
\label{tab:absolute}
\setlength{\tabcolsep}{3.4pt}
\begin{tabular}{llrrrr}
\toprule
Data & Metric & Q3 & Q7 & I2 & I8 \\
\midrule
\multirow{2}{*}{VSR} & $C$ & 14.33 & 5.93 & 9.73 & 13.36 \\
                     & $R$ &  6.54 & 4.87 & 5.66 &  6.05 \\
\multirow{2}{*}{GQA} & $C$ & 12.15 & 1.91 & 3.86 &  5.55 \\
                     & $R$ &  6.23 & 3.74 & 4.30 &  5.43 \\
\bottomrule
\end{tabular}
\end{table}

Coverage gains should be read together with the operating point itself.
Table~\ref{tab:absolute} reports Relation-Orbit's mean asserted fraction and
mean realized selective risk. All eight mean risks are below the calibration
target of $10\%$, although individual splits can exceed it and coverage varies
materially by model and dataset. Counting abstain-all splits as satisfying the
target under the zero-risk convention above, the fraction of test splits
meeting the target ranges from $76.3\%$ to $85.7\%$. In particular, GQA with Qwen-7B asserts only
$1.91\%$ of claims. Relation-Orbit therefore improves the usable region under
this calibration protocol; it does not make every left--right claim safe to
assert. Figure~\ref{fig:gqa-case} illustrates one GQA split in which Relation-Orbit
meets its held-out threshold while the one-sided score has no feasible
nonempty acceptance threshold.

\subsection{Structure diagnostic and limitations}

We separately test the group-level partition under a two-sided
selective-classification diagnostic: $\hat y=\mathbf 1[s\ge0]$, reliability
$u=|s|$, and acceptance $u\ge\tau$, using 300 grouped $50/50$
calibration/test splits. This is not the assertion protocol of
Tables~\ref{tab:main}--\ref{tab:absolute}. Across 12 dataset--backbone settings on VSR, GQA, and COCO, the correct orbit exceeds the mean and median of the 34 label-independent alternative
partitions and the mean of the 16 single-swap corruptions in all 12 settings.
Its average rank is $1.61/35$, and it lies in the top quartile in every
setting. Some strongest post-hoc alternatives still perform better, supporting
a systematic structural advantage rather than unique optimality. Partition orientation is fixed without labels. Moreover, a label-independent
one-claim-per-image robustness check preserves rank $1/35$ in all eight
VSR/GQA settings, indicating that the ranking is not driven by counting both
members of each inverse-claim pair.

The 34 alternatives hold the observation count fixed and test alternative
four-versus-four organizations. Each of the 16 corruptions exchanges one
supporting observation with one counterfactual observation. All controls are
label-independent, so this diagnostic audits structure rather than fitting a
partition. It remains separate because two-sided reliability and one-sided
assertion are different tasks.

In the main protocol, scores below the calibration-selected threshold lead to
abstention; the sign alone is not correctness, and the score is not a probability.

Relation-Orbit requires likelihood access and eight view--formulation scores,
excluding text-only APIs and increasing compute. It also relies on a valid
transformation: text, handedness, or other flip-sensitive cues can invalidate
reflection, and domain shift requires transformation checks and
recalibration~\cite{ovadia2019shift}. An above--below pilot with vertical reflection produced no stable
matched-budget gains, underscoring that a semantically valid visual
involution cannot be assumed across relations; stringent targets may yield abstain-all.

\section{Conclusion}
\label{sec:conclusion}

Relation-Orbit is a parameter-free contrast for selective left--right claim
verification with frozen VLMs. By organizing reflection-induced likelihoods
according to inverse-relation and entity-exchange roles, it improves coverage
over one-sided and Orbit-Max controls and remains favorable to reduced-orbit
alternatives in most settings under the evaluated held-out calibration
protocols. A separate label-independent partition diagnostic supports a
systematic, but not uniquely optimal, structural advantage.

\clearpage
\bibliographystyle{IEEEbib}
\bibliography{references}

\end{document}